\documentclass[sigconf,screen]{acmart}

\usepackage{lipsum}
\usepackage{url}
\usepackage{float}
\usepackage{placeins}
\usepackage{balance}

\AtBeginDocument{%
  \providecommand\BibTeX{{%
    \normalfont B\kern-0.5em{\scshape i\kern-0.25em b}\kern-0.8em\TeX}}}

\copyrightyear{2022}
\acmYear{2022}
\setcopyright{acmlicensed}\acmConference[MM '22]{Proceedings of the 30th ACM International Conference on Multimedia}{October 10--14, 2022}{Lisboa, Portugal}
\acmBooktitle{Proceedings of the 30th ACM International Conference on Multimedia (MM '22), October 10--14, 2022, Lisboa, Portugal}
\acmPrice{15.00}
\acmDOI{10.1145/3503161.3548101}
\acmISBN{978-1-4503-9203-7/22/10}

\begin{document}

\title{Talking Head from Speech Audio using a Pre-trained Image Generator}


\author{Mohammed M. Alghamdi}
\affiliation{
  \institution{University of Leeds}
%   \city{Leeds}
%   \country{The United Kingdom}
}
\affiliation{
  \institution{Taif University}
%   \city{Taif}
%   \country{Saudi Arabia}
}
\email{scmmalg@leeds.ac.uk}

\author{He Wang}
\affiliation{%
  \institution{University of Leeds}
%   \city{Leeds}
%   \country{The United Kingdom}
}
\email{h.e.wang@leeds.ac.uk}

\author{Andrew J. Bulpitt}
\affiliation{%
  \institution{University of Leeds}
%   \city{Leeds}
%   \country{The United Kingdom}
}
\email{a.j.bulpitt@leeds.ac.uk}

\author{David C. Hogg}
\affiliation{%
  \institution{University of Leeds}
%   \city{Leeds}
%   \country{The United Kingdom}
}
\email{d.c.hogg@leeds.ac.uk}


\begin{abstract}
    We propose a novel method for generating high-resolution videos of talking-heads from speech audio and a single 'identity' image. Our method is based on a convolutional neural network model that incorporates a pre-trained StyleGAN generator. We model each frame as a point in the latent space of StyleGAN so that a video corresponds to a  trajectory through the latent space. Training the network is in two stages. The first stage is to model trajectories in the latent space conditioned on speech utterances. To do this, we use an existing encoder to invert the generator, mapping from each video frame into the latent space. We train a recurrent neural network to map from speech utterances to displacements in the latent space of the image generator. These displacements are relative to the back-projection into the latent space of an identity image chosen from the individuals depicted in the training dataset. In the second stage, we improve the visual quality of the generated videos by tuning the image generator on a single image or a short video of any chosen identity. We evaluate our model on standard measures (PSNR, SSIM, FID and LMD) and show that it significantly outperforms recent state-of-the-art methods on one of two commonly used datasets and gives comparable performance on the other. Finally, we report on ablation experiments that validate the components of the model. The code and videos from experiments can be found at \url{https://mohammedalghamdi.github.io/talking-heads-acm-mm/} 
\end{abstract}





\begin{teaserfigure}
  \includegraphics[width=\textwidth]{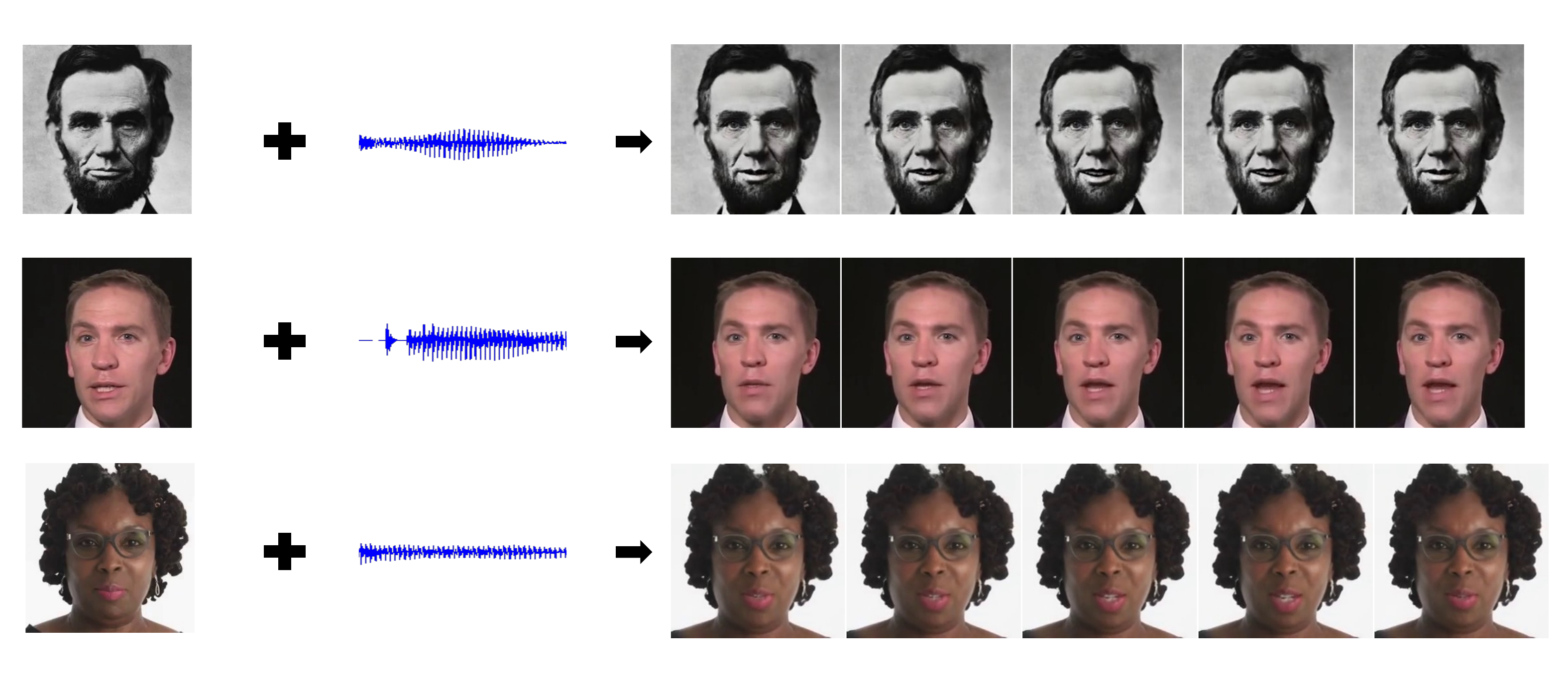}
  \caption{Given a single identity image and speech audio, our model generates high-resolution talking-head video of the identity lip-synced with the audio.}
  \Description{NOTE: TBC}
  \label{fig:teaser}
\end{teaserfigure}

\maketitle

\section{Introduction}

Synthesising video of talking heads from speech audio has many potential applications, such as video conferencing, video animation production and virtual assistants. Although there has been considerable prior work on this task, the quality of generated videos is typically limited in terms of overall realism and resolution. In this paper, we propose an audio-driven model that synthesises high-resolution talking-head videos (1024 x 1024 in our experiments) from a single identity image.  \\

Many previous models generate low-resolution video \cite{Vougioukas2018EndtoEndSF,Vougioukas2019RealisticSF} or cropped faces \cite{Chung17b, chen2019hierarchical}. Low resolution video is generally not suitable for deployment in many real world applications, such as a virtual assistant. A common approach has been to use intermediate features such as facial landmarks to map from audio to output video \cite{Yang2020MakeItTalk, chen2019hierarchical,Das2020SpeechDrivenFA}.  Another approach has been to edit an existing talking-head video to blend in a new mouth region synthesised from audio  \cite{thies2020nvp, Suwajanakorn2017}.  \\

Recent advances in image synthesis have been successful at generating high-resolution images from noise \cite{stylegan1,karras2017progressive, stylegan3}. Karras et al. \cite{stylegan1} propose a style-based generator StyleGAN that synthesises high quality images that are largely indistinguishable from real ones. Some works have studied the latent space of StyleGAN \cite{roich_pivotal_2021, abdal2019image2stylegan, shen2020interpreting, harkonen2020ganspace} and discovered meaningful semantics for manipulating images. Recent work has leveraged the richness of a pre-trained StyleGAN generator \cite{karras2020stylegan2} to generate high-resolution videos from noise by decomposing (disentangling) the motion and content in the latent space \cite{tian_good_2021, fox_stylevideogan_2021, skorokhodov_stylegan-v_2021}. Tian et al. \cite{tian_good_2021} discover motion trajectories in the latent space to render high-resolution videos while image and motion generators are trained on different domain datasets. \\

Inspired by these advances, we propose a novel method for generating high-resolution videos conditioned on speech audio by constructing trajectories in the latent space of a pre-trained image generator \cite{karras2020stylegan2}. Our framework uses a pre-trained image encoder \cite{richardson2021encoding} to find the latent code of a given identity image in the latent space of the generator. We then train a recurrent audio encoder along with a latent decoder to predict a sequence of latent displacements to the encoded identity image. In this stage, we show our approach can generate talking-head videos with accurate mouth movements conditioned on speech audio. To improve the visual quality of the generated videos further, we tune the generator on a single image or short video of a target subject using the PTI \cite{roich_pivotal_2021} method. We compare our approach with other state of the art approaches qualitatively and quantitatively using benchmark measures: LMD, SSIM, PSNR and FID. We show that it achieves performance at least as good as the state of the art on two commonly used datasets. \\

Our principal contributions are:

\begin{itemize}
  \item A method for generating high-resolution videos from speech audio by constructing motion trajectories in the latent space of a pre-trained image generator;
  \item A comparative evaluation, including a user study, demonstrating the performance of the method on both quantitative and qualitative criteria.
  
  %%our model using a set of evaluation measures with previous state of the art works. We conduct user studies to evaluate the visual quality and lip-synchronisation of the generated videos. We show that our method outperforms state of the art methods. 
\end{itemize}

\begin{figure*}[tp]
  \includegraphics[width=\textwidth]{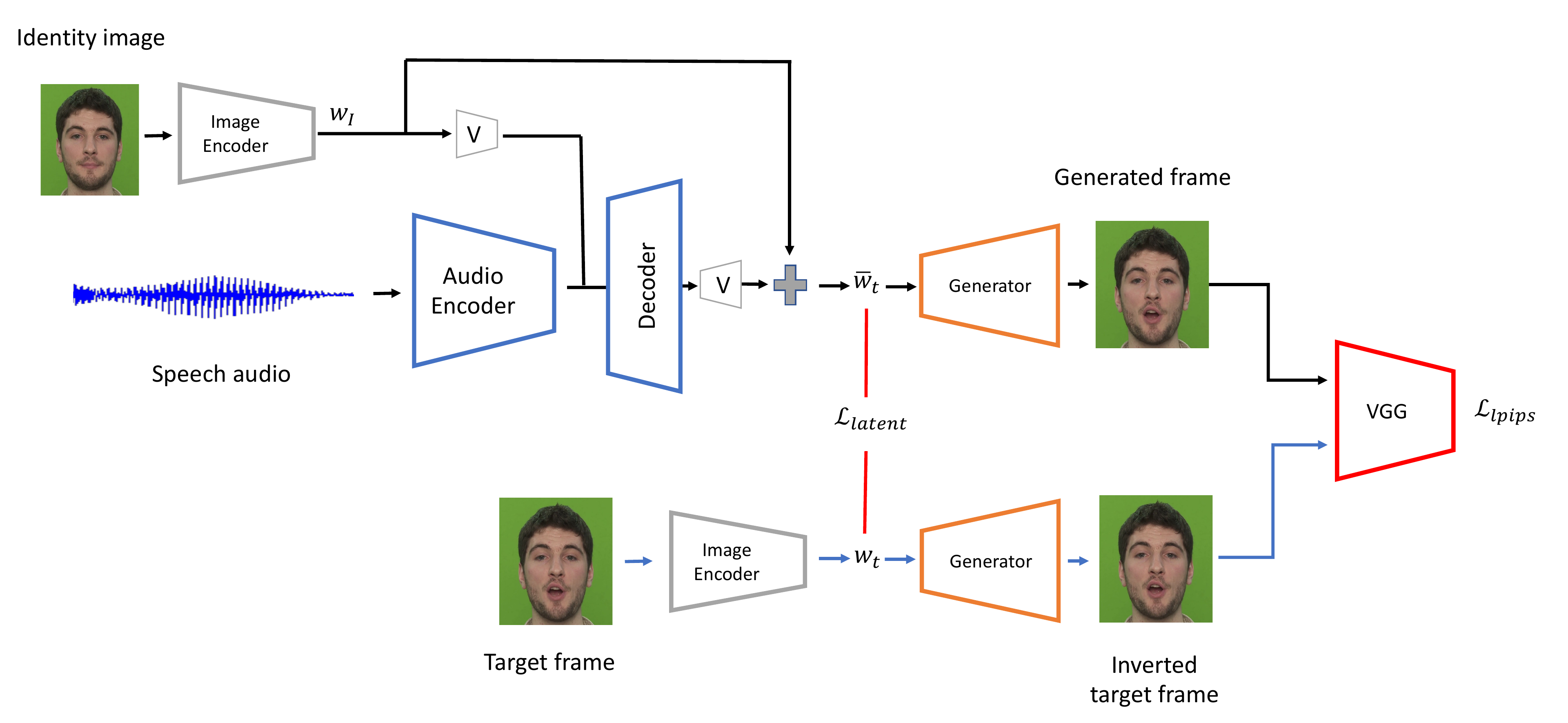}
  \caption{Overview of the model: Given an identity image and a speech audio, the aim is to synthesise a video of the identity lip-synced with the audio. We first find the corresponding latent code of the identity image using the image encoder $E_I$. We then encode the audio using the audio encoder $E_A$. Next, we embed the identity latent code using the PCA basis $V$. The decoder $D$ then takes both the embedded identity latent vector and encoded audio to predict a displacement to the identity latent vector in the latent space of a pre-trained image generator $G$.}
  \label{fig:model}
\end{figure*}

\section{Related Work}

\subsection{Audio-driven talking-head generation}
Various methods have been proposed to generate videos of talking heads. Given audio, the task is to lip-sync the head to the audio. The audio may itself be generated automatically from text or be extracted from a video clip of someone speaking. Some of these methods are generic, and can generate videos of any identity given one or more images of that identity ~\cite{Chung17b,Vougioukas2018EndtoEndSF,Vougioukas2019RealisticSF,chen2019hierarchical,zhou2018talking,zhou2021pose,Prajwal2020Wave2Lip,Yang2020MakeItTalk}. However, they can only generate low-resolution videos. Other methods that generate high-resolution videos from speech audio ~\cite{thies2020nvp, Suwajanakorn2017, Meshry_2021_ICCV,Zhang_2021_ICCV,ji2021audio-driven, linsen2020ebt, Audio2head, zhang2021flow, Lahiri_2021_CVPR} can only work on a single subject. These approaches require retraining the models for each new subject. \\

Chung et al. \cite{Chung17b} propose a model to generate videos using multiple images of the target face and an audio speech sample. The model consists of an audio encoder and identity encoder that learn a joint embedding of the face and audio, and a decoder that generates a frame that best represents the audio sample for the target identity. Prajwal et al. \cite{Prajwal2020Wave2Lip} adopt a similar approach, except that a pretrained lip-sync discriminator, and a visual discriminator are used in addition to $L_1$ loss. Vougioukas et al. \cite{Vougioukas2018EndtoEndSF} expand the approach by training a recurrent-based decoder with a noise generator to model spontaneous facial expressions (e.g. blinks). 
Other models rely on 2D intermediate features (e.g. facial landmarks) to learn the mapping between audio input and video output~\cite{chen2019hierarchical,Yang2020MakeItTalk, Das2020SpeechDrivenFA}.  Chen et al. propose a cascade approach that generates a talking face video given an image and audio. The method first transfers the given audio signal to facial landmarks and then generates video frames conditioned on the landmarks.  Zhou et al. \cite{Yang2020MakeItTalk} adopt a similar approach that first disentangles the content and speaker information from the input audio. Then, these two components are mapped to content and speaker facial landmark spaces using a recurrent model on each. \\

Other methods rely on 3d intermediate features (e.g. through monocular reconstruction) to synthesise high-quality videos of a single subject.  Some attempt to generate only the mouth region and blend it to a target video ~\cite{Suwajanakorn2017, thies2020nvp, linsen2020ebt, Lahiri_2021_CVPR}. Thies et al. \cite{thies2020nvp} propose an approach that predicts the coefficients that drive a person-specific expression blendshape basis using audio features. A neural texture rendering network is then used to generate the mouth region. 
In addition, others modify the facial expressions, geometry or pose of a target video conditioned on the  audio~\cite{Zhang_2021_ICCV, ji2021audio-driven}. In contrast, our approach generates a full-frame video of a talking-head without editing a target video or relying on intermediate features. 

% \subsection{latent space directions}
% - manipulating images in the latent space 

\subsection{Unconditional video generation using StyleGAN}
Recently, there have been several works that use a pre-trained image generator (StyleGAN) to generate videos  \cite{skorokhodov_stylegan-v_2021,fox_stylevideogan_2021,tian_good_2021}. They all share the idea of discovering motion trajectories in the latent space of a StyleGAN generator without conditioning on any driving source. Tian et al. \cite{tian_good_2021} propose a MOCOGAN-HD model that uses a motion generator to predict residual latent codes from an initial latent code sampled from the latent space of StyleGAN. The model is trained in the image space with a multi-scale video discriminator as well as contrastive image discriminator. Similarly, Fox et al. \cite{fox_stylevideogan_2021} reduce the training cost by training a Wasserstein GAN model in the latent space instead of image space. Although their model is trained on a single subject dataset, it can transfer the learned motion to a new subject using an offset trick. Skorokhodov et al. \cite{skorokhodov_stylegan-v_2021} modify the StyleGAN network to learn a continuous latent trajectory using a neural representation based approach. Our work differs from these methods by learning the motion trajectories in the latent space of StyleGAN conditioned on speech audio.

\vspace{10 pt}

\section{The Method}

Our method consists of four components: image encoder $E_I$, audio encoder $E_A$, latent decoder $D$ and image generator $G$. Given an identity image I and speech audio $\mathbf{a}$, the goal is to synthesise a video of the identity lip-synced with the audio. We first partition the audio clip into a sequence of $T$ fixed-duration audio segments $\{ a_1, a_2,..,a_T \}$. From this audio segment sequence, we target a video clip consisting of a sequence of $T$ video frames $\{ x_1, x_2,..,x_T \}$. There is therefore a one-to-one correspondence between input audio segments and output video frames.

The inference pipeline is as follows. We take the identity image I and finds its latent code $w_I$ in the latent space $W^{+}$ of the generator $G$  using the image encoder $E_I$. Next, we encode an audio segment $a_t$ using the audio encoder $E_A$ and extract an audio embedding vector $e_t = E_A(a_t)$. We then feed in the identity latent code $w_I$ and the audio vector $e_t$ jointly to the latent decoder $D$ to predict a latent displacement $d_t = D(w_I,e_t)$. We then calculate the displaced latent vector $\Bar{w}_t = d_t + w_I$ . Lastly, the image generator $G$ takes the displaced latent vector $\Bar{w}_t$ and generates the corresponding video frame $\Bar{x}_t$. An overview of the model can be seen in figure \ref{fig:model}. In the following section, we describe each component in detail.

\subsection{Architecture}

\subsubsection{Image generator}
For our image generator $G$, we use the pre-trained StyleGAN \cite{karras2020stylegan2} trained on the FFHQ dataset \cite{stylegan1} to synthesise static images of faces. The StyleGAN architecture consists of mapping and synthesis networks. The mapping network is a non-linear 8-layer MLP which maps a latent code $z$ sampled from a latent space $Z$ to an intermediate space $W^{+}$. The produced $w$ controls the synthesis network through an adaptive instance normalization (AdaIN) operation after each convolutional layer. We use only the synthesis part as our image generator $G$.

\subsubsection{Image encoder}
\begin{figure}
  \includegraphics[width=\linewidth]{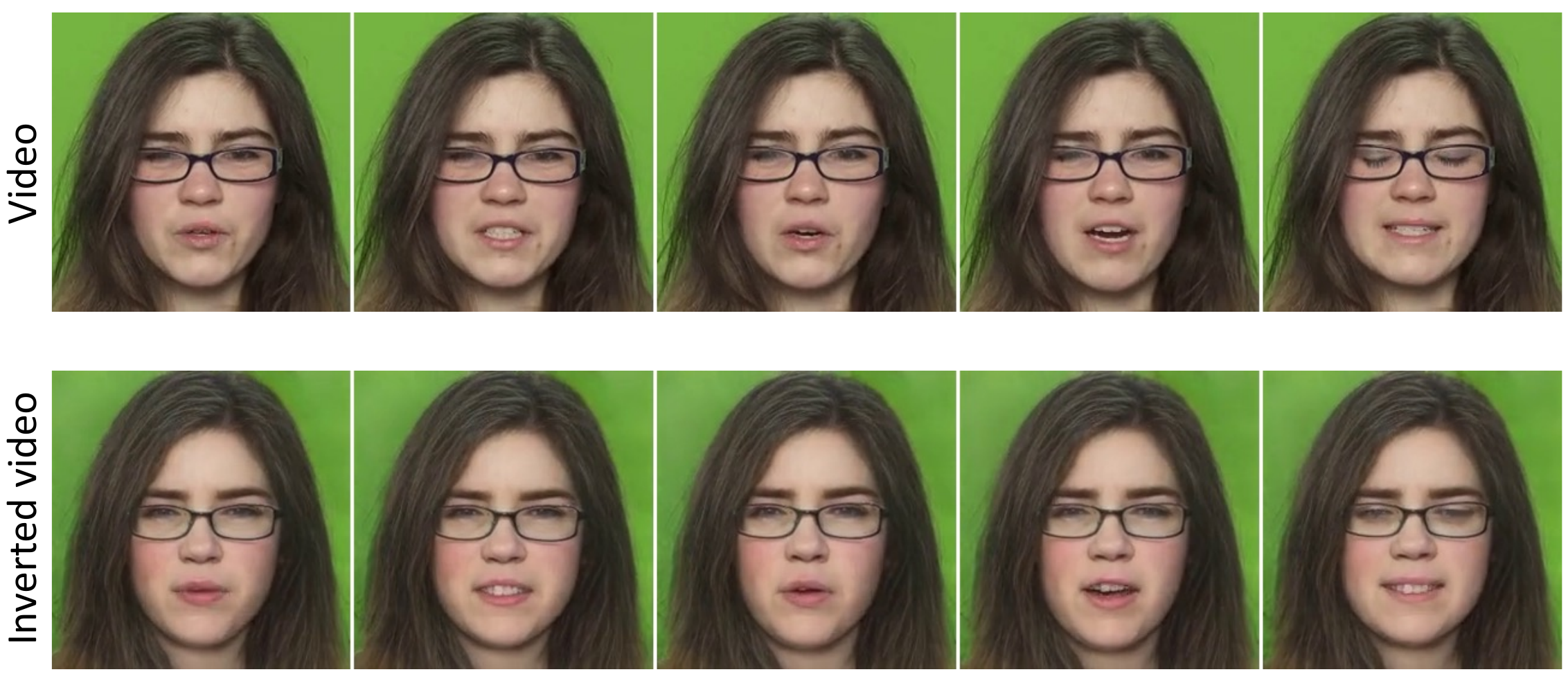}
  \caption{This figure illustrates how the image encoder $E_I$ can encode a video (top row) into the latent space of StyleGAN $W^{+}$  while preserving its mouth movements and other facial expressions. Bottom row shows the inverted video.}
  \label{fig:how_enc_preserves}
\end{figure}

We use pSp \cite{richardson2021encoding}, an off-the-shelf pre-trained image encoder that inverts real images to the $W^{+}$ space of StyleGAN. The encoder was trained by embedding the FFHQ \cite{stylegan1} dataset to a fixed StyleGAN generator. Critically, the encoder produces latent vectors that preserve the mouth expression on synthesis with StyleGAN. Figure \ref{fig:how_enc_preserves} shows frames from a video that has been inverted into the latent space $W^{+}$ using the image encoder $E$ and then re-generated using $G$. 

%  learning directions on the speech audio
\subsubsection{Audio encoder}
We represent each audio segment $a_t$ using Mel-frequency cepstral coefficients (MFCC). The audio encoder $E_A$ is a recurrent model that takes a sequence of audio segments $\{ a_1, a_2,..,a_T \}$ and produces a sequence of encoded audio segments  $\{ e_1, e_2,..,e_T \}$. The encoder network $E_A$ consists of multiple convolutional layers followed by three LSTM layers.  

\subsubsection{Latent decoder}
The latent decoder $D$ takes as input the identity latent vector $w_I$ and an encoded audio segment $e_t$ to predict a displacement $d_t$ to the identity vector in the latent space $W^{+}$. To reduce the high dimensionality of the latent space $W^{+}$, we first conduct principal component analysis (PCA) on the FFHQ dataset \cite{stylegan1} mapped into $W^{+}$ using the image decoder. We obtain a subspace from the components with the largest eigenvalues, giving a basis $\mathbf{V}$. We project the identity latent $w_I$ input into the subspace defined by $\mathbf{V}$ and concatenate it with the encoded audio segment $e_t$. This provides the input to the decoder. We map the decoder's output $h_t$ from the subspace to $W^{+}$ to get the displacement vector $d_t$. Thus, we obtain $\Bar{w_t}$ as follows:

\begin{equation}\label{eqn:decoder}
    \Bar{w_t} = w_I+d_t = w_I + h_t \cdot \mathbf{V}, ~~~ t = 1, 2, 3, \cdots, T,
\end{equation}

\subsection{Training}
Our model is trained in two stages. In stage one, we are only interested in learning trajectories in the latent space $W^{+}$ conditioned on the speech audio. The model predicts latent displacements to the identity in the latent space of a fixed image generator. This disentangles the mouth motion and the image content. The motion trajectories are learned by training the model using a talking-head dataset. Although the model learns to generate accurate mouth movements, the visual quality of the generated video exhibits some distortion (see section \ref{sec:ablation}). The quality is determined by the pre-trained StyleGAN generator, which has been trained on images of people who are typically making a static pose. In stage two, we tune the generator $G$ on a single image or short video of a target speaker.  

\subsubsection{Stage one} % Learning latent speech directions
We train only the audio encoder $E_A$ and latent decoder $D$ while keeping the pre-trained image encoder $E_I$ and the pre-trained image generator $G$ fixed. For the loss function, we project the target frame $x_t$ into the generator's latent space $W^{+}$ using the image encoder $E_I$. We then have the corresponding latent code $w_t$ = $E_I(x_t)$ for $t = 1,2, ... , T$. We calculate an L2 loss between each target latent code $w_t$ and the predicted latent code $\Bar{w}_t$. We define $\mathcal{L}_\mathrm{latent}$ as follow:

\begin{equation}\label{eqn:latent_loss}
\begin{gathered}
 \mathcal{L}_\mathrm{latent} = \sum_{t=1}^{T} 
 || w_t - \Bar{w}_t ||_2  \\
\end{gathered}
\end{equation}

In addition, we apply another loss in the image domain between the generated video $\mathbf{\Bar{x}}$ and the target video $\mathbf{x}$. Since the pre-trained image generator has not seen the training data, applying the loss directly on the target video would affect the model's performance, enforcing it to focus on the facial appearance rather than mouth movements. For this, we invert the target video $\mathbf{x}$ using the image encoder $\hat{x_t} = E_I(x_t)$. We calculate the  perceptual loss \cite{zhang2018perceptual} between the generated video $\{ \Bar{x}_1, \Bar{x}_2,..,\Bar{x}_T \}$ and the inverted target video  $\{ \hat{x}_1, \hat{x}_2,..,\hat{x}_T \}$ as follows: 

\begin{equation}\label{eqn:lpips_loss}
\begin{gathered}
 \mathcal{L}_\mathrm{LPIPS} = \sum_{t=1}^{T} 
 ||\phi(\Bar{x_t}) - \phi(G_I(E_I(x_t))) ||_2  \\
\end{gathered}
\end{equation}

where $\phi$ is based on the VGG neural network \cite{vgg}. Thus, the overall loss for learning to predict latent displacements driven by audio is a weighted sum of the two losses:

\begin{equation}\label{eqn:stage1_loss}
\begin{gathered}
  \mathcal{L}_\mathrm{stage1} =   \lambda_\mathrm{latent}\mathcal{L}_\mathrm{latent} + \lambda_\mathrm{LPIPS}\mathcal{L}_\mathrm{LPIPS}\\
\end{gathered}
\end{equation}

\begin{figure*}[ht!]
  \includegraphics[width=\textwidth]{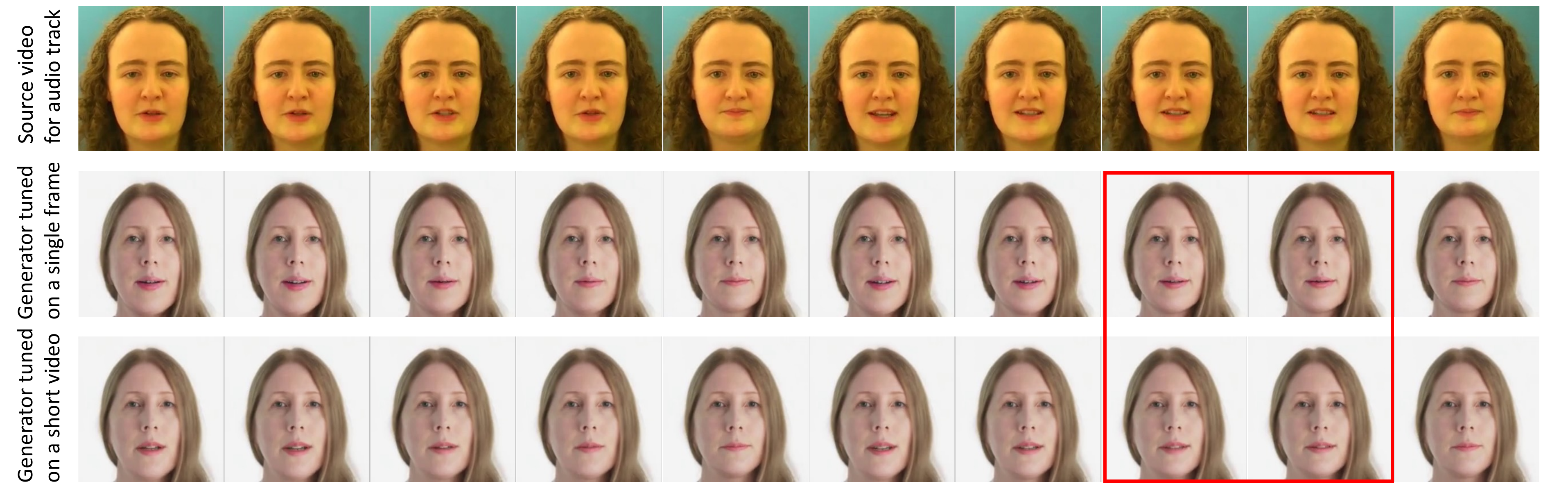}
  \caption{Samples generated using our approach. The top row shows frames from a source video providing the audio used to drive the generation. The middle row shows the corresponding generated frames where the generator $G$ is tuned on a single frame. The bottom row shows generated frames where the generator $G$ is tuned on a 5-second video clip. These videos are included in the supplementary material.}
  \label{fig:our_comp}
\end{figure*}

\subsubsection{Stage two} One could tune both the image encoder $E_I$ and the image generator $G$ on the target speaker using an auto-encoder. However, this would transform the latent space $W^{+}$ and the learned model at stage one would consequently fail to generate correct mouth movements. To improve the visual quality of the generated video, we use the PTI method \cite{roich_pivotal_2021} to tune only the generator on a single image or video of a target speaker. In experiments, we implement stage two on short videos, and as a limiting case, on a single image.  \\

Given a video $\mathbf{x}$ of a speaker, we tune the image generator $G$ on the video frames $x_t$. For this task, we use the pre-trained encoder $E_I$ to encode the frames $x_t$ to the latent space $W^{+}$ to get $w_t$. Given $\hat{x}_t = G(w_t; \theta^*)$, we tune the weights of the generator while keeping the encoder fixed. We use the same objective loss used in PTI \cite{roich_pivotal_2021}: 

\begin{equation}\label{eqn:pti_loss}
\begin{gathered}
\mathcal{L}_\mathrm{stage2} =  \mathcal{L}_\mathrm{LPIPS} + \mathcal{L}_\mathrm{L2}
\end{gathered}
\end{equation}

% \begin{equation}\label{eqn:pti_loss}
% \begin{gathered}
% \mathcal{L}_\mathrm{stage2} =  \mathcal{L}_\mathrm{LPIPS}(x_t, \hat{x}_t) + \mathcal{L}_\mathrm{L2}(x_t, \hat{x}_t)
% \end{gathered}
% \end{equation}

where $\mathcal{L}_\mathrm{L2}$ is defined as :
% squared error loss defined as $(x_t - \hat{x}_t)^2$.
\begin{equation}\label{eqn:l2_loss}
\begin{gathered}
 \mathcal{L}_\mathrm{L2} = \sum_{t=1}^{T} 
 ||(x_t - \hat{x}_t ||_2  \\
\end{gathered}
\end{equation}

After tuning the generator, we can generate videos of talking-heads using our inference pipeline with the components trained in both stages.

\section{Experiments}
\subsection{Datasets}
We evaluate our approach using two widely used datasets for synthesizing talking-head videos: GRID \cite{GRID} and TCD-TIMIT \cite{TCDTIMIT}. The GRID dataset has 33 speakers uttering 1000 short sentences each containing 6 words. The TCD-TIMIT has 59 speakers each uttering 100 sentences. We hold-out ten speakers from each dataset for testing and use the remaining for training. The videos are resampled to 25 fps. To align the video frames, we use the same face alignment method used in preprocessing the FFHQ dataset \cite{stylegan1} for training the original StyleGAN \cite{karras2020stylegan2}. The input to the audio encoder $E_A$ is an audio segment of length 0.2 seconds which corresponds to a window of five frames. However, we choose only the middle frame as the the ground truth frame. The identity image is a randomly chosen frame out of this window. We represent the audio speech using MFCC values extracted from the raw values. Each audio segment is a window of size 12 x 28, where the columns represent MFCC features for each time step.    

\subsection{Implementation details}
We perform our experiments using PyTorch \cite{pytorch}. For the image generator, we use an unofficial implementation of StyleGAN\footnote{https://github.com/rosinality/stylegan2-pytorch}. For the pre-trained image encoder, we use the official implementation of p2p \cite{richardson2021encoding}. For training the audio encoder $E_A$ and the decoder $D$ in stage one, we use an Adam optimiser \cite{kingma2014adam} with a learning rate of 0.0002. In Eqn. \ref{eqn:stage1_loss}, we set $\lambda_\mathrm{latent}$ = 250 and $\lambda_\mathrm{LPIPS}$ = 1. We tune the generator with a learning rate of 0.0003. The tuning process takes less than two minutes for a single identity image. All experiments use an NVIDIA V100 GPU with 32 GB of memory.

% \subsection{Qualitative results of stage 1}
\begin{figure*}[tp]
  \includegraphics[width=\textwidth]{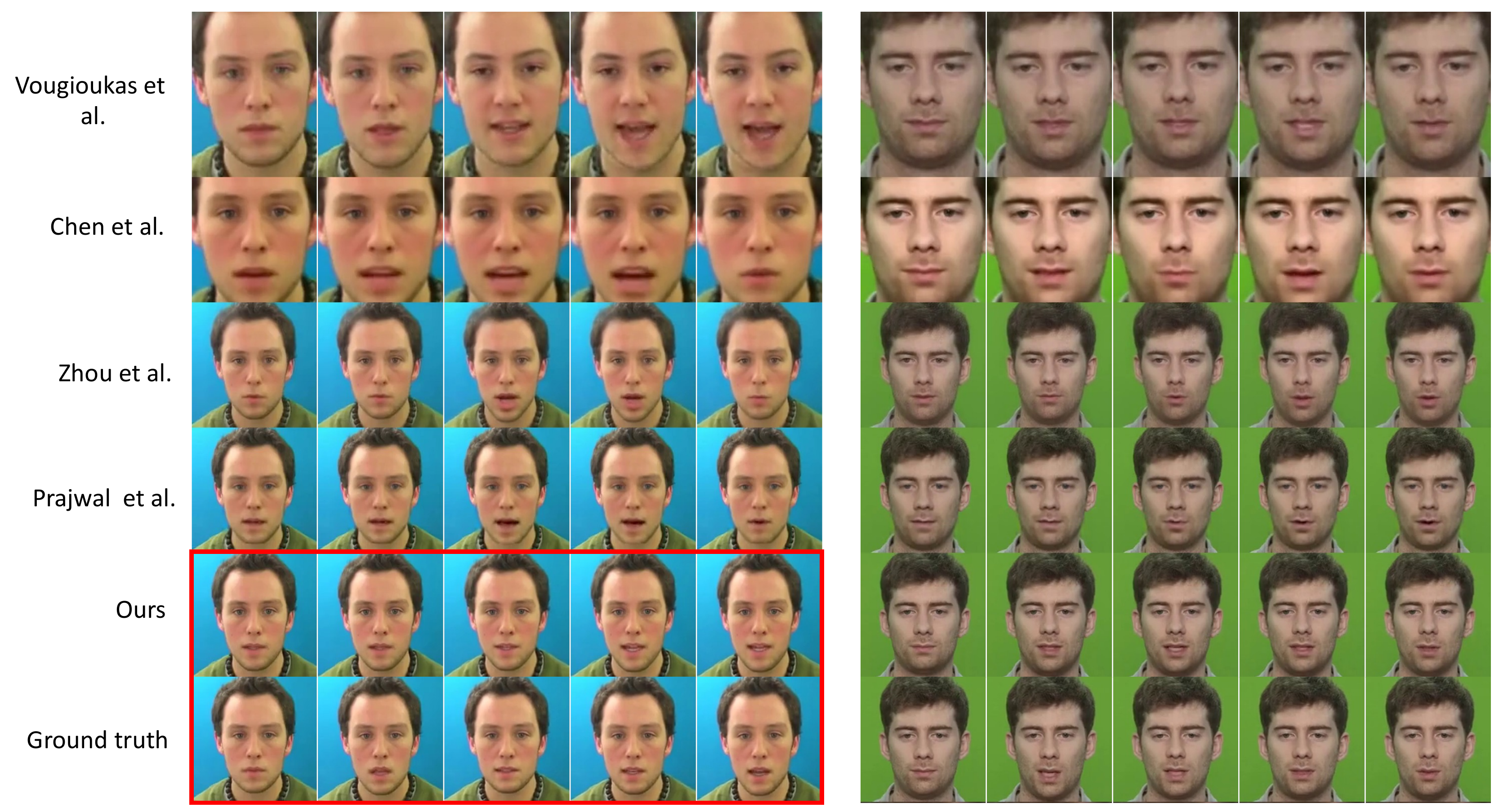}
  \caption{Qualitative comparisons. The videos are generated by the methods of Vougioukas et al. \cite{Vougioukas2018EndtoEndSF}, Chen et al. \cite{chen2019hierarchical}, Zhou et al. \cite{Yang2020MakeItTalk}, Prajwal et al. \cite{Prajwal2020Wave2Lip} and \textit{Ours} on audio samples form TCDTIMIT (left) and GRID (right). It can be seen that our model synthesises a lip-movement that is closer to the ground truth than the other methods.}
  \label{fig:comp_image}
\end{figure*}

\begin{table*}
  \caption{We conducted quantitative comparisons in two benchmark datasets.}
    \label{tab:quan_eval}
    \centering
  \begin{tabular*}{\linewidth}{p{4.1cm}p{1.2cm}p{1.2cm}p{1.2cm}p{1.2cm} | p{1.2cm}p{1.2cm}p{1.2cm}p{1.2cm}}
      \toprule
      \toprule
Method & \multicolumn{4}{c}{TCD-TIMIT} & \multicolumn{4}{c}{GRID}   \\
      \midrule
      
&{PSNR$\uparrow$}&{SSIM$\uparrow$}&{FID$\downarrow$}&{LMD$\downarrow$}
&{PSNR$\uparrow$}&{SSIM$\uparrow$}&{FID$\downarrow$}&{LMD$\downarrow$} \\
\hline
{Vougioukas, et al. ~\cite{Vougioukas2018EndtoEndSF}} 
&{17.24}&{0.60}&{16.05}&{3.42}   
&{16.72}&{0.62}&{13.58}&{3.08}    \\ 
\hline

{Chen, et al. ~\cite{chen2019hierarchical}} 
&{15.31}&{0.58}&{11.79}&{3.66} 
&{16.80}&\textbf{{0.69}}&{13.27}&{3.74}   \\ 
\hline
{Zhou, et al. ~\cite{Yang2020MakeItTalk}} 
&{18.10}&{0.58}&{18.02}&{2.59} 
&{18.53}&{0.61}&{11.87}&{2.64}     \\ 
\hline
{Prajwal, et al.~\cite{Prajwal2020Wave2Lip}} 
&{18.26}&{0.64}&{15.24}&{2.19} 
&{17.83}&\textbf{{0.69}}&{11.11}&\textbf{{2.05}}    \\

  \hline
 {\textit{Ours}}  
&{\textbf{20.55}}&\textbf{{0.65}}&\textbf{{8.11}}&\textbf{{2.18}}
&\textbf{{20.33}}&{0.65}&\textbf{{5.30}}&{2.18}      \\ 
 \hline

      \bottomrule
  \end{tabular*}

\end{table*}

\subsection{Results}
In this section, we evaluate our model after tuning the generator. Figure \ref{fig:our_comp} shows the quality of the generated videos from tuning the generator on a single frame (middle row) and on a short video (bottom row). The figure shows that tuning the generator on multiple frames has resulted in better visual quality. This can be seen in the mouth appearance highlighted in red.

To evaluate the quality of the generated videos, we use two common reconstruction measures: The peak signal-to-noise ratio (PSNR) and the structural similarity (SSIM) \cite{psnr_ssim}. For these measures, a larger score is better. We use a landmarks distance metric (LMD) \cite{LMD} to evaluate the synchronisation between the mouth movement and the speech audio. This metric computes the Euclidean distance between mouth landmarks of each generated frame and its corresponding true frame. It then averages the score on the number of frames and number of mouth landmark points. We also use a Fréchet Inception Distance (FID) to quantitatively evaluate generated videos. For LMD and FID measures, a lower score is better. 

We compare our work against four state of the art models \cite{Vougioukas2018EndtoEndSF,chen2019hierarchical,Yang2020MakeItTalk, Prajwal2020Wave2Lip} and use the official available codes of these models to generate the videos and compute the evaluation measures. Table \ref{tab:quan_eval} shows that our approach outperforms other state of the art models on the TCD-TIMIT dataset \cite{TCDTIMIT}. On the GRID dataset \cite{GRID}, the model achieves better scores on the PSNR and FID measures. Figure \ref{fig:comp_image} shows the visual quality of generated videos by our model in comparison with other models. The highlighted frames show that our model generates photo-realistic videos largely indistinguishable from the ground truth. 

\begin{figure*}[tp]
  \includegraphics[width=\textwidth]{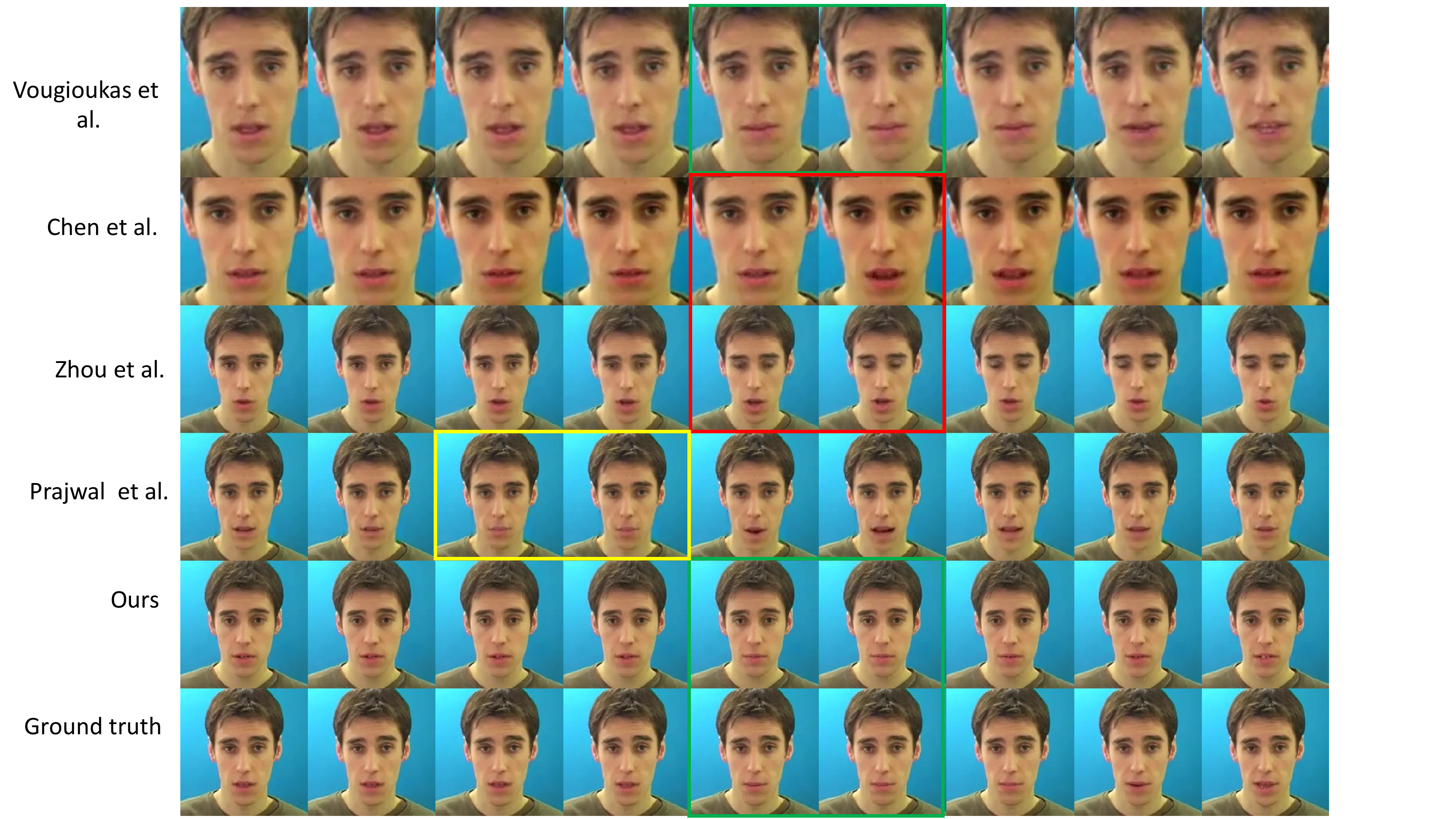}
  \caption{Visual comparison of mouth-closure during bilabial events. The green highlighted areas show closed lip gestures generated by ours and Vougioukas et al in comparison with the ground truth. The yellow area shows out of sync mouth-closure generated by Prajwal et al. while the red area shows failure in producing a closed mouth for Chen et al. and Zhou et al. The video is included in the supplementary material.}
  \label{fig:bilabial}
\end{figure*}

Figure \ref{fig:bilabial} shows a visual comparison of our model with others on a challenging mouth movement associated with the phoneme /p/ in the word "place".  It can be seen that our model and Vougioukas et al.  produce a closed-mouth shape (highlighted in green) in sync with the ground truth  while  Prajwal et al. (highlighted in yellow) is out of sync with the ground truth. In addition, Chen et al. and Zhou et al. (highlighted in red) fail to produce the required mouth shape.

Table \ref{tab:compexity} shows the number of trainable parameters, inference time and output frame size for the five methods. We ran all experiments on a V100 Nvidia GPU and report the achieved frame rate (FPS) as a measure of inference time. The source videos are sampled at 25 FPS. We can see that the method of Vougioukas, et al. \cite{Vougioukas2018EndtoEndSF} and ours are faster than real-time. In addition, our method generates much higher resolution videos compared to others.

\begin{table}[b]
\caption{Comparisons between our method and others in terms of number of parameters, inference time and output size. }
\label{tab:compexity}
\centering
\begin{tabular}{p{2.8cm}p{1.55cm}p{1.33cm}p{1.33cm}}
  \toprule
  \toprule
 Method & Number of parameters  & Inference time & Output size \\ 
  \midrule
 Vougioukas, et al. \cite{Vougioukas2018EndtoEndSF} & 55.28 M & 441 FPS & 96x128 \\ 
 \hline
 Chen et al. \cite{chen2019hierarchical} & 88.43 M & 15.57 FPS & 128x128 \\
 \hline
 Zhou, et al. \cite{Yang2020MakeItTalk} & 36.40 M & 10.32 FPS & 256x256 \\
 \hline
 Prajwal, et al. \cite{Prajwal2020Wave2Lip} & 36.30 M & 16 FPS & 256x256 \\
 \hline
 Ours & 29.68 M & 35 FPS & 1024x1024 \\
 \hline
\end{tabular}
\end{table}

\subsection{User Study}

\begin{figure}[b]
  \includegraphics[width=\linewidth]{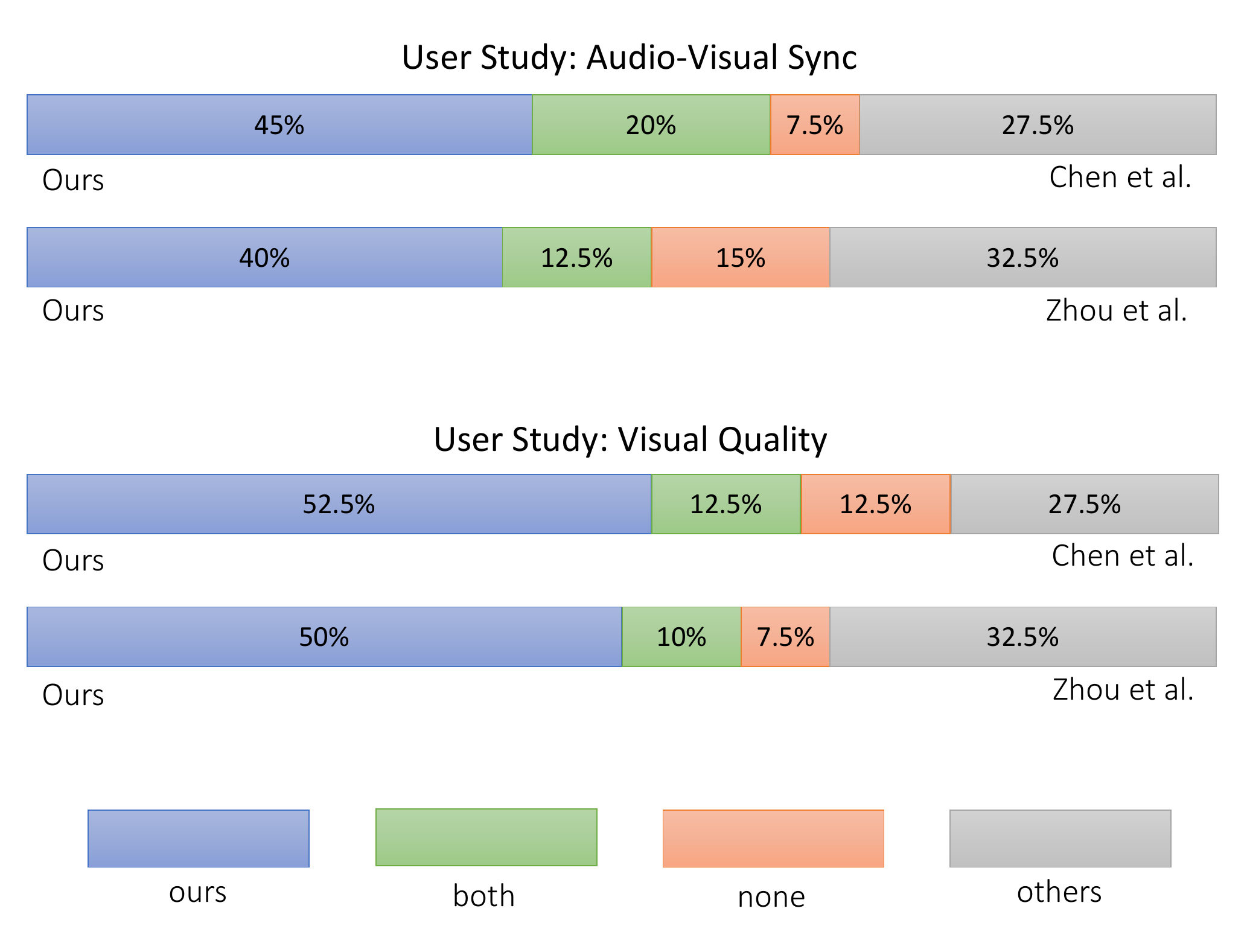}
  \caption{User study results for evaluating our approach with state of the art methods in terms of the audio-visual synchronisation (top) and and the visual quality (bottom)}
  \label{fig:user_study}
\end{figure}

We conducted a user study to compare our approach with related works using Amazon Mechanical Turk services.
We evaluate both the audio-visual synchronisation and the visual quality of the state-of-the-art methods. We show participants a pair of videos: one generated by our method and the other generated by either Chen et al. \cite{chen2019hierarchical} or Zhou et al. \cite{Yang2020MakeItTalk}. For each pair, we either ask which video looks more photo-realistic or which video has more accurate lip-sync with the audio. The choices for each question are "right", "left", "none", and "both". We randomly choose the order of videos in each pair. We obtained 80 answers from 20 participants for each question. Figure \ref{fig:user_study} shows the results of the user study. It can be seen our model achieves a better result in terms of both the audio-visual sync and the visual quality.

\subsection{Ablation analysis}
\label{sec:ablation}
We analyse the effect of each loss in Eqn \ref{eqn:stage1_loss} on the performance of the model in generating talking-head videos. We train the model in stage one without $\mathcal{L}_\mathrm{latent}$ and $\mathcal{L}_\mathrm{LPIPS}$ separately and test it after tuning the generator in stage two. We observe that the choices of these losses do not affect the visual quality of the generated video but affect the lip-synchronisation accuracy. This is indicated in table \ref{tab:ablation_study_1} on the LMD column. The model trained on both losses outperforms the model trained on either of the losses alone. \\

\begin{figure}[b]
  \includegraphics[width=\linewidth]{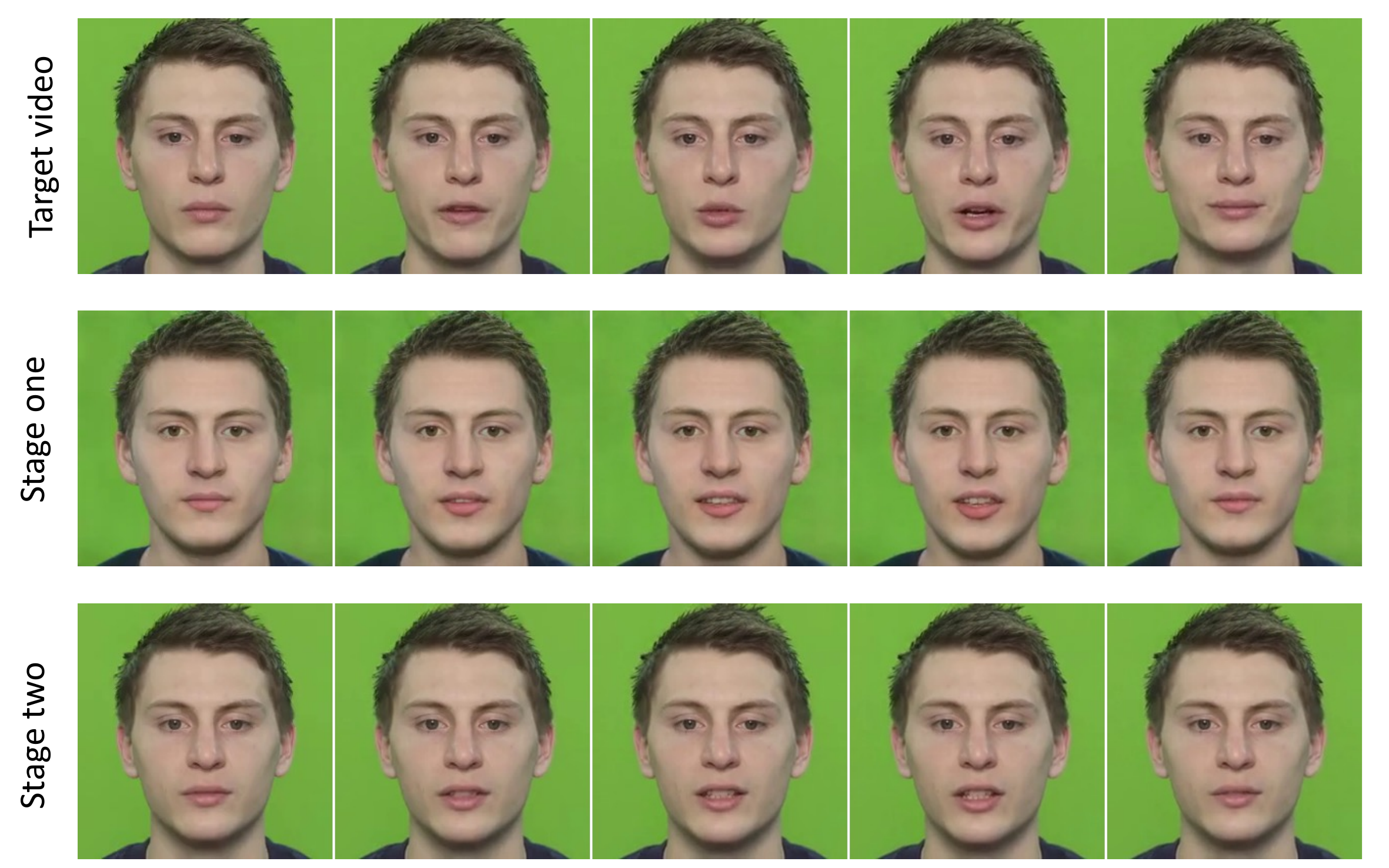}
  \caption{This figure compares the visual quality between a video generated before tuning the generator $G$ (stage one) and after tuning the generator (stage two). The top row is the target video.}
  \label{fig:ablation_figure}
\end{figure}

We also compare the performance of the model at stage one without tuning the generator and after tuning the generator. Figure \ref{fig:ablation_figure} shows a sample of generated video using the model trained in stage one. It can be seen that the model generates correct mouth positions but the visual quality inherits some distortion caused by the image generator. In stage two, we tune the generator on a single frame or short video of the target speaker. We can see from Table \ref{tab:ablation_study_2} that SSIM and PSNR are higher after tuning the generator, indicating that stage two is important to improve the quality of the generated video.

\begin{table}[b]
  \caption{Ablation analysis on losses in Eqn \ref{eqn:stage1_loss} }
    \label{tab:ablation_study_1}
    \centering
  \begin{tabular}{p{3.0cm}|p{1.0cm}p{1.0cm}p{1.0cm}}
      \toprule
      \toprule
Method  & \multicolumn{3}{c}{TCD-TIMIT}  \\
      \midrule
      
&{PSNR$\uparrow$}&{SSIM$\uparrow$}&{LMD$\downarrow$}  \\
\hline
 w/o $\mathcal{L}_\mathrm{latent}$, in Eqn \ref{eqn:stage1_loss}
&{20.57}&{0.65}&{2.30}\\ 
  \hline
w/o $\mathcal{L}_\mathrm{LPIPS}$, in Eqn \ref{eqn:stage1_loss}
&\textbf{{20.78}}&\textbf{{0.66}}&{2.75}\\ 
\hline
 {\textit{Proposed Model}}  
&{20.55}&{0.65}&\textbf{{2.18}}\\ 
 \hline
      \bottomrule
  \end{tabular}

\end{table}

\begin{table}[b]
  \caption{Comparisons between the performance of the model before and after tuning the generator.}
    \label{tab:ablation_study_2}
    \centering
  \begin{tabular}{p{3.0cm}|p{1.0cm}p{1.0cm}p{1.0cm}}
      \toprule
      \toprule
Method  & \multicolumn{3}{c}{TCD-TIMIT}  \\
      \midrule
      
&{PSNR$\uparrow$}&{SSIM$\uparrow$}&{LMD$\downarrow$}  \\
\hline
 {Stage one only} 
&{17.55}&{0.49}&{2.37}\\ 
  \hline
 {\textit{Proposed Model}}  
&\textbf{{20.55}}&\textbf{{0.65}}&\textbf{{2.18}}\\ 
 \hline
      \bottomrule
  \end{tabular}

\end{table}

\section{Ethical consideration}
Our framework can synthesise high quality videos from speech audio. This is perfect for video production animation, a talking-head avatar and video-dubbing. Creative people may use our work to edit content in movies or generate new videos. However, the model can be misused to spread misinformation or manipulate existing data. Generated videos using deep learning are becoming hard to distinguish from the real thing, although there have been promising advances in forensics on the detection of  "deepfake" videos \cite{rossler2019faceforensics++,cozzolino2018forensictransfer}. We will share generated videos of our framework with the community to help detecting manipulated videos. 

\section{Conclusion and future works}
We propose a novel method for synthesising high-resolution videos from speech audio. The model can generate videos of a target speaker given a short video (or single image) of the speaker. Our model is built on top of a pre-trained image generator. We first learn to generate talking-head videos by constructing motion trajectories conditioned on speech audio. We then improve the image generator by tuning it on a short video of a target speaker. 

We show that the method significantly outperforms recent state-of-the-art methods on TCD-TIMIT in quantitative experiments and gives performance comparable to the state-of-the art on GRID. The method also performs best in the user study.

The generated faces depict only mouth movements because the training datasets (TCD-TIMIT and GRID) are neutral and expressionless. We anticipate our approach could in principle generate other facial expressions where these are present in the dataset (e.g. \cite{mead}), but have not yet demonstrated that this is the case.

%%%%  old sentences  %%%%
% Although our approach can generate accurate lip movements, it cannot not model spontaneous expressions e.g. eye blinks. The reason could be that some of these expressions are correlated with the audio. In future work, we plan to control other facial expressions in the latent space of StyleGAN. 
%%%% 

\begin{acks}
We are grateful to Rebecca Stone and Jose Martinez for their comments on this paper. This work was undertaken on ARC4, part of the High Performance Computing facilities at the University of Leeds. MA is supported by a PhD scholarship from Taif University.
\end{acks}

\FloatBarrier
\bibliographystyle{ACM-Reference-Format}
\balance
\bibliography{sample-base}

% \appendix


\end{document}